\documentclass[9pt,shortpaper,twoside,web]{ieeecolor}
\usepackage{cite}
\usepackage{amsmath,amssymb,amsfonts}
\usepackage{xspace}
\usepackage{graphicx}
\usepackage{hyperref}
\usepackage{textcomp}
\usepackage[ruled, vlined]{algorithm2e}
\def\BibTeX{{\rm B\kern-.05em{\sc i\kern-.025em b}\kern-.08em
    T\kern-.1667em\lower.7ex\hbox{E}\kern-.125emX}}
\begin{document}
\title{SplitAVG: A heterogeneity-aware federated deep learning method for medical imaging}
\author{Miao Zhang, Liangqiong Qu, Praveer Singh, Jayashree Kalpathy-Cramer, Daniel L. Rubin
\thanks{Manuscript received June 20, 2021. This work was supported in part by a grant from the NCI, U01CA242879. (corresponding authors: Liangqiong Qu and Daniel L. Rubin}
\thanks{Miao Zhang and Liangqiong Qu are with Department of Biomedical Data Science, Stanford University, United States (e-mail: miaozhangmi@gmail.com; liangqi@stanford.edu).}
\thanks{Praveer Singh and Jayashree Kalpathy-Cramer are Athinoula A. Martinos Center for Biomedical Imaging, Massachusetts General Hospital, United States.  (e-mail: psingh19@mgh.harvard.edu; kalpathy@nmr.mgh.harvard.edu).}
\thanks{Daniel L. Rubin is with Department of Biomedical Data Science and Radiology, Stanford University, United States (e-mail: dlrubin@stanford.edu).}
\thanks{Miao Zhang and Liangqiong Qu contributed equally.}}

\maketitle

\begin{abstract}
Federated learning is an emerging research paradigm for enabling collaboratively training deep learning models without sharing patient data. However, the data from different institutions are usually heterogeneous across institutions, which may reduce the performance of models trained using federated learning. In this study, we propose a novel heterogeneity-aware federated learning method, SplitAVG, to overcome the performance drops from data heterogeneity in federated learning. Unlike previous federated methods that require complex heuristic training or hyper parameter tuning, our SplitAVG leverages the simple network split and feature map concatenation strategies to encourage the federated model training an unbiased estimator of the target data distribution. We compare SplitAVG with seven state-of-the-art federated learning methods, using centrally hosted training data as the baseline on a suite of both synthetic and real-world federated datasets. We find that the performance of models trained using all the comparison federated learning methods degraded significantly with the increasing degrees of data heterogeneity. In contrast, SplitAVG method achieves comparable results to the baseline method under all heterogeneous settings, that it achieves 96.2\% of the accuracy and 110.4\% of the mean absolute error obtained by the baseline in a diabetic retinopathy binary classification dataset and a bone age prediction dataset, respectively, on highly heterogeneous data partitions. We conclude that SplitAVG method can effectively overcome the performance drops from variability in data distributions across institutions. Experimental results also show that SplitAVG can be adapted to different base convolutional neural networks (CNNs) and generalized to various types of medical imaging tasks. The code is publicly available at \href{https://github.com/zm17943/SplitAVG}{https://github.com/zm17943/SplitAVG}.
\end{abstract}


\begin{figure*}[!t]
\centerline{\includegraphics[width=\columnwidth]{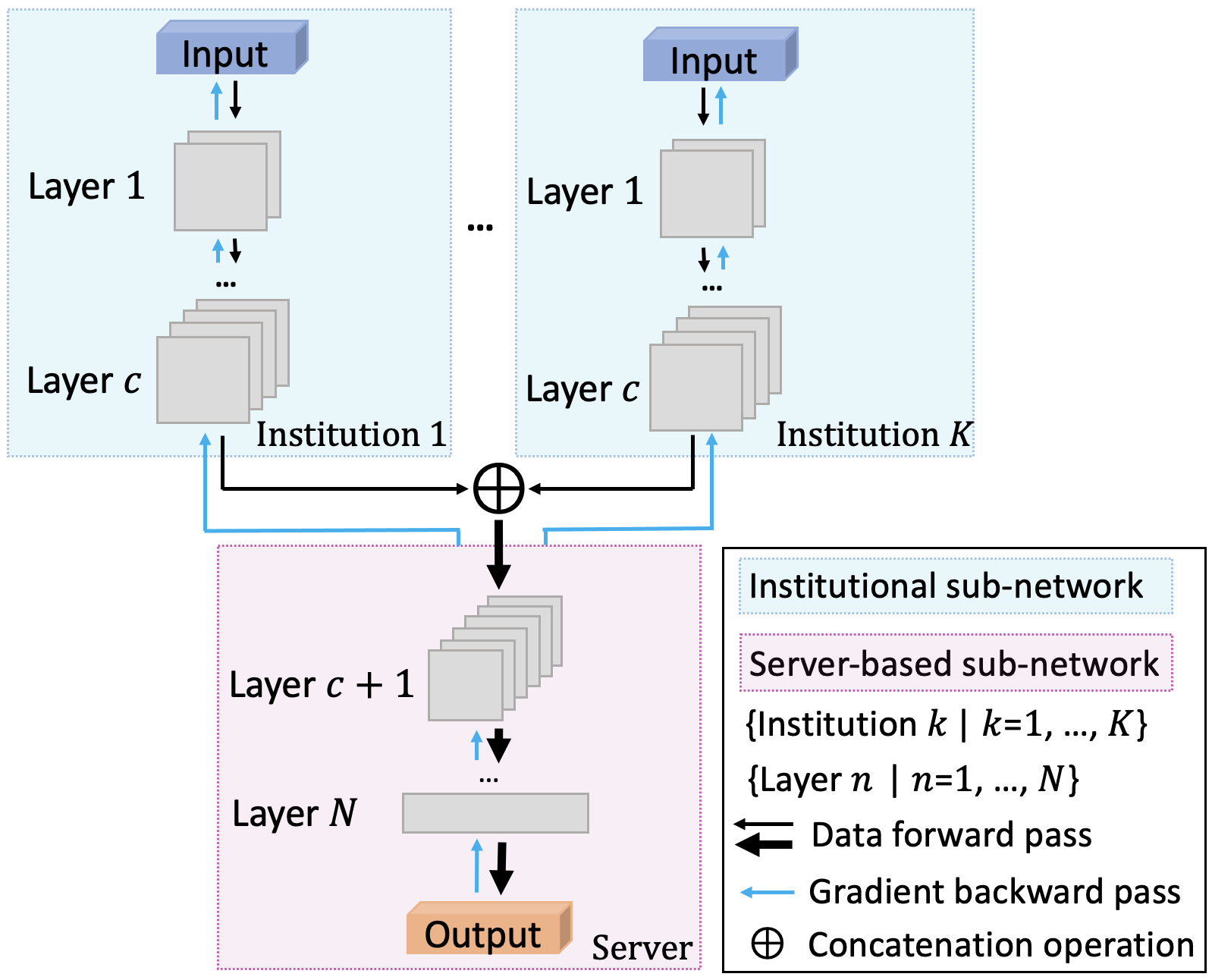}}
\caption{Architecture of Split Averaging (SplitAVG): A deep learning network is split into two sub-networks at a pre-defined cut layer. An institutional sub-network resides at each local institution, and a server-based sub-network resides on a central server.}
\label{Architecture}
\end{figure*}
\section{Introduction}
\label{sec:introduction}
Deep learning techniques and advances in computer hardware offer the promise of great advances in various medical applications, e.g., diagnosing ocular diseases \cite{Zhao2019-ne}, detecting glaucoma from optical coherence tomography images \cite{Wang2019-qs}, identifying serious illnesses with natural language processing \cite{Udelsman2019-yq}, and providing appropriate treatment recommendation \cite{Hwang2019-mr}. However, training a robust deep learning model that generalizes across centers often requires a tremendous amount of training cases. The amount of patient data at individual medical institutions, or even in public data repositories such as The Cancer Imaging Archive, is often limited, especially for rarer diseases \cite{Tresp2016-gc, Clark2013-ty}. Aggregating patient data from multiple centers is often complicated owing to patient privacy, legal regulatory barriers to data sharing, and inefficiency of moving large amounts of data. As such, federated learning (also termed as “collaborative learning” or “distributed learning”), where the training of a global deep learning model is performed locally at each institution without sharing raw data, has become a promising alternative for accessing large scale data to train robust deep learning models \cite{Shokri2015-xu,  Jochems2016-si, McMahan2017-ip}.

Existing federated learning methods can be grouped into aggregation-based federated learning methods \cite{McMahan2017-ip} and transfer-based methods \cite{Chang2018-rg, Gupta2018-dq}. Aggregation-based federated learning methods repeatedly average weights or gradient updates of the models trained at local institutions, such as Federated Averaging (FedAvg) and Federated stochastic gradient descent (FedSGD) \cite{McMahan2017-ip}. Transfer-based methods train a model at each local institution for specific number of iterations and then transfer full or part of the model weights to next institution until model convergence, such as in Cyclical weight transfer (CWT) \cite{Chang2018-rg} and SplitNN \cite{Gupta2018-dq} respectively.

One of the key challenges to federated learning algorithms is mitigating the effects of data heterogeneity among participating institutions on performance of the final learned model \cite{Fernandez2018-zp}. The distributed nature of federated learning means that there can be substantial heterogeneity in the distributions of training data across institutions. Aggregation-based methods like FedAvg can be robust to certain non-IID (independent and identically distributed) settings \cite{McMahan2017-ip}, but the accuracy of the synchronized averaged model reduces significantly on highly skewed data partitions \cite{Zhao2018-kq}. Transfer-based methods may also lose model performance when heterogeneity exists in the training data across institutions, since the model trained in cyclic transferring way always suffers from catastrophic forgetting on non-IID settings \cite{Lee2017-rk}. It has been demonstrated that training on non-IID data partitions is a pervasive problem for federated learning methods, and it always degrades the performance of deep learning models \cite{Hsieh2020-wx}.

Several recent efforts have been devoted to overcoming the degrading effects of heterogeneous data across institutions in federated learning. The related studies include adding momentum to server model weight updates to prevent client updates trained on non-IID data partitions from diverging (FedAvgM) \cite{Hsu2019-sd}, applying the group normalization (Group Norm) \cite{Wu2020-sh} layers as the alternative of batch normalization to avoid the skew-induced accuracy loss of the batch normalization layer for non-IID data (FedSGD+GD) \cite{Hsieh2020-wx, Valiant1990-ts}, and sharing a subset of global data with local institutions to improve the training of FedAvg by mitigating weight divergence due to non-IID data (FedAVG+SD) \cite{Zhao2018-kq}. Though these are promising approaches, the current approaches only work well on partitions with mild data distribution heterogeneity and still suffer performance drops on highly heterogeneous cases (see the comparison results in Fig.5 for details).

In this study, we propose a novel federated learning technique, Split Averaging (SplitAVG), to overcome the deleterious effects of heterogenous data distributions across institutions\footnote{The data heterogeneity in our study indicates heterogeneity from cross-institution data, also written as non-IID or data skew.} in federated learning. At the heart of SplitAVG is a network splitting operation and an intermediate feature map concatenation strategy. Specifically, our SplitAVG splits the network into an institutional sub-network (residing at the local institutions) and a server-based sub-network (residing on a central server) at a predefined layer of the network (see Fig.~\ref{Architecture}). As the training examples in each local institution are sampled from institution-specific data distribution, which is a biased-estimator of the actual distribution of the whole population on non-IID data partitions, we further apply a concatenation operation on the central server to concatenate all the intermediate feature maps collecting from the institutional sub-networks. This concatenation operation allows our SplitAVG method to learn from the union of institution-specific data distribution rather than directly learning a biased-estimator of the actual distribution of the whole population, thus working well on both IID and non-IID data partitions. Our experimental results demonstrate the capability of the proposed method in handling unbalanced and non-IID data partitions.

The remainder of this paper is organized as follows: 1) present our SplitAVG algorithm, detail its forward propagation and back propagation training stage, 2) detail the binary classification dataset and regression datasets used to evaluate our method, 3) compare SplitAVG with seven state-of-the-art federated learning methods and the baseline centrally-hosted method, 4) present the experimental setup, and 5) provide an experimental evaluation of SplitAVG and its comparison methods on both IID data partitions and various non-IID data partitions.

\section{MATERIALS and METHODS}
\subsection{SplitAVG}
In this section, we outline our proposed federated learning platform, SplitAVG (see Fig.~\ref{Architecture}), and provide a prospective to understand the advantage of the proposed method.

Define the deep learning network involved in SplitAVG as a function $F$, which consists of a list of $N$ sequential layers, i.e., $F=\{l1,l2,…,lc,…,lN\}$. In SplitAVG, we split F into two sub-networks at a specific layer (also known as cut layer) $lc$ and rewrite  $F=\{FI,FS\}$, where $FI=\{l1,l2,…,lc\}$ is the institutional sub-network that resides at the local institutions, and $FS=\{l(c+1),l(c+2),…,lN\}$ is the server-based sub-network that resides on a central server. In each round of federated training, each local institution trains the institutional sub-network $FI$ in parallel with its local data, sends the output feature maps to the central server for concatenating with those from other institutions, then the server completes the rest of the training with the aggregated feature maps on the server-based sub-network. Specifically, as depicted in Algorithm \ref{algo:b}, SplitAVG follows a two-stage training phase: 1) data forward propagation procedure from institutional sub-network to the server-based sub-network with the transfer of the intermediate feature maps and their corresponding labels, 2) aggregated gradient back propagation procedure from server-based sub-network to the institutional sub-network. This two-stage training process is continued until model convergence on a separate validation set or the maximum number of training epochs is reached. Once the training is finished, the server will send the weights of server-based sub-network $FS$ back to each local institution. Then each institution can perform validation and testing with the complete network $F=\{FI,FS\}$.
\begin{algorithm*}

\newcommand{\vect}{\ensuremath{\mathbf{##1}}}

\textbf{Initialize:}   \\
$F = \{ FI,FS\} $  \\ $\triangleright$ Institutions and the server split $F$ into institutional sub-network $FI$ and server-based sub-network $FS$  \\

$ FI = \{ l1, l2, …, lc \} $

$ FS = \{l(c+1), l(c+2), …, lN\}$ \\ $\triangleright$ $lc$ as the cut layer  \\

Server initializes weight $W$ for $FI$ and sends $W$ to $K$ institutions \\

Server initializes weight $W_S$ for $FS$ \\

\text{\break}

\textbf{for} each epoch i from $1$ to $E$ \textbf{do}: \\

\quad Server samples a subset of $St$ institutions \\

\quad \textbf{for} each institution $k \in St$ \textbf{in parallel do}:

\quad \quad For each local batch ${x_k,y_k}$ in ${\boldsymbol{x}_k,\boldsymbol{y}_k}$ do:

\quad \quad \quad \textbf{Forward Propagation:} \\

\quad \quad \quad $x_k^{lc} \leftarrow FI_k(x_k) $   \\
\quad \quad \quad $\triangleright$ Institution $k$ forward propagates $x_k$ to cut layer $lc$  \\

\quad \quad \quad Server $\leftarrow$ $x_k^{lc}, y_k$ \\
\quad \quad \quad $\triangleright$ Institution $k$ sends intermediate feature maps at $lc$ and labels to server \\

\quad \quad \quad $X_S^{lc} \leftarrow {x_1^{lc}  \oplus x_2^{lc}… \oplus x_{St}^{lc}} $ \\
\quad \quad \quad $\triangleright$ Server concatenates feature maps from all institutions

\quad \quad \quad $Y_S \leftarrow {y_1 \oplus y_2… \oplus y_{St}}  $ \\
\quad \quad \quad $\triangleright$ Server concatenates labels from all institutions \\

\quad \quad \quad $\hat{Y}_S \leftarrow FS(X_S^{lc}) $ \\

\quad \quad \quad $\triangleright$ Server forward propagates concatenated feature maps $X_S^{lc}$ till the final layer

\text{\break}

\quad \quad \quad \textbf{Back propagation:} \\

\quad \quad \quad $g^{lN} = \bigtriangledown  \mathcal{L} (\hat{Y}_S,Y_S)$   \\
\quad \quad \quad $\triangleright$ Server generates gradients at output layer

\quad \quad \quad $g^l \leftarrow {W_S^{l+1}}^T g^{l+1}, \text{for} \text{\quad}  l=l(N-1),l(N-2),…,l(c+1)  $ \\
\quad \quad \quad $\triangleright$ Server back propagates gradients to $l(c+1)$ \\

\quad \quad \quad Institution $k \in St \leftarrow g^{l(c+1)}$ \\
\quad \quad \quad $\triangleright$ Server sends gradients at $l(c+1)$ to each local institution \\

\quad \quad \quad $g^l \leftarrow {W_k^{l+1}}^T g^{l+1},\text{for} \text{\quad} l=lc,l(c-1),…,l1 $ \\
\quad \quad \quad $\triangleright$ Institution $k$ back propagates gradients \\

\quad \quad \quad  $W_k^l \leftarrow W_k^l- \eta g^l, \text{for} \text{\quad} l=lc,l(c-1),…,l1$ \\
\quad \quad \quad $\triangleright$ Institution $k$ updates $FI$

\quad \quad \quad $W_S^l \leftarrow W_S^l- \eta g^l, \text{for} \text{\quad} l=lN,l(N-1),…,l(c+1) $ \\
\quad \quad \quad $\triangleright$ Server updates $FS$

\quad \textbf{Weight Transfer}   \\
\quad Institution $k \in {1,…,K} \leftarrow W_S $ \\
\quad $\triangleright$ Server sends $FS$ to complete institution models

\text{\break}
 \caption{SplitAVG. The $K$ institutions are indexed by $k$; $E$ is the number of training epochs; $\{x_k,y_k\}$ is the data at institution $k$; $F=\{l1,l2,…,lc,...,lN\}$ is the network function, and $l$ is a layer in $F$. $lc$ is the cut layer. $\eta$ is the learning rate.}
\label{algo:b}
\end{algorithm*}

\subsubsection{Forward Propagation} Let’s assume there are total $K$ local institutions involved in federated learning, indexed by $k$, and denote the training data of institution $k$ as  $\{ \boldsymbol{x_k}, \boldsymbol{y_k} \}$. In SplitAVG, we select a subset of $St \ll K$ local institutions following the client-selecting methods in FedAvg ~\cite{McMahan2017-ip}, and then start the following forward propagation steps: 1) apply standard forward propagation on institutional sub-network $FI$ with sampled min-batch $\{x_k,y_k\}$ in each selected local institution, getting intermediate feature maps $FI(x_k)$; 2) send the intermediate feature maps and their corresponding labels $\{FI(x_k),y_k\}$ to the central server; 3) concatenate the received feature maps $X_S^{lc}= \{FI_1 (x_1)\oplus {FI}_2 (x_2)...\oplus {FI}_{St} (x_{St})\}$ and their corresponding labels $Y_S= \{y_1\oplus y_2...\oplus y_{St}\}$ at the central server; and finally 4) forward propagate the combined feature maps into the server-based sub-network $FS(X_S^{lc})$. This will complete a round of forward propagation without sharing the raw data. Unlike traditional federated learning methods, such as FedAvg \cite{McMahan2017-ip}, that directly average the model weights learned from institutional specific data distribution, where the synchronized averaged central model will lose accuracy or even completely diverge when high heterogeneity exists in the data partitions across sites\cite{Zhao2018-kq}, our concatenation of feature maps on the central server guarantees that the server-based sub-network is trained on the union of all the institutional data and not from biased institution-data, thus it works well on both IID and non-IID data partitions.

\subsubsection{Back Propagation} After a round of forward propagation, SplitAVG back propagates the gradients from the last layer of the server-based sub-network to the first layer of the institutional sub-network $FI$. Given the loss function $\mathcal{L}$, the detailed back propagation of SplitAVG is shown as the following: 1) calculate the gradients $g^{lN}=\bigtriangledown \mathcal{L} (Y_S,{FS}(X_S^{lc}))$ of the server-based sub-network $FS$;  2) back propagate gradients $g^{lN}$ at the central server from the last layer of server-based sub-network $FS$ to its first layer, and denote the gradient at the first layer of server-based sub-network as $g^{l(c+1)}$ ; 3) transfer the gradients $g^{l(c+1)}$  back to each local institution and complete the rest of the back-propagation operation through each institutional sub-network $FI$; 4) update the model weights of both the server-based sub-network $FS$ and the institutional sub-network $FI$. Our back propagation procedure strictly follows the chain rule in differentiation, and it will achieve exactly the same results as the normal deep learning training procedure.

As opposed to traditional federated learning methods (such as FedAvg and FedSGD) \cite{McMahan2017-ip} that require frequent transfers of model weights or model gradients of the entire network, in SplitAVG, only the intermediate feature maps $X_S^{lc}$  and gradients  $g^{l(c+1)}$ at the cut layer are communicated between local institutions and the central server, which greatly reduces the computation and communication costs.  In addition, the direct feature map concatenation step in the central server provides convergence guarantees for the model training and allows us to train a robust model on both IID and non-IID data partitions.

\subsection{Theory analysis for SplitAVG}
SplitAVG improves local models with biased data generalizing to global data distributions. To illustrate, we use the performance bound for aggregated federated learning models built in previous work \cite{zhu2021data}: Assuming there are $K$ local institutions. $T_K$ is the $k^{th}$ institution’s task domain where the raw data distribution is $D_k$, the local model learnt on this domain is $h_k$, and the empirical risk of the model on $T_k$ is ${\hat{\mathcal{L}}}_{T_k}(h_k)$. $T$ is the global domain where the data distribution $D$ including $m$ samples is assumed to be unbiased. $h$ is a global model achieved from federated learning which aims to minimize the risk of task from all $K$ local site, written as $\mathcal{L}_T (h)$. $\mathcal{L}_T (h)$ has an upper bound that with probability larger than $1-\delta$:
\begin{align*}
  \mathcal{L}_T(h) \equiv \mathcal{L}_T (\frac{1}{K}\sum_k h_k)
  \\
  \leq \frac{1}{K}\sum_k {\hat{\mathcal{L}}}_{T_k}(h_k) + \frac{1}{K}\sum_k (d({\tilde{D}_k}, {\tilde{D}}) + \lambda_k) 
  \\
  +\sqrt{\frac{4}{m}(dlog\frac{2em}{d} + log\frac{4K}{\delta})}
\end{align*}

$d$ denotes the divergence measured between two domains, and $\tilde{D}_k$ and $\tilde{D}$ are intermediate feature representations reduced from raw images in $D_k$ and $D$ by a same feature extraction structure. The implication was derived that data heterogeneity of local sites to the global distribution leads to a high representation divergence $d(\tilde{D}_k, \tilde{D})$. The divergence increases the risk bound of the aggregated model $h$ on the global domain thus diminishes model quality. SplitAVG was motivated to reduce feature representation divergence, by proposing concatenation operation on representations from selected local client: $\tilde{D}_k \leftarrow  \{\tilde{D}_1 \oplus \tilde{D}_2...\oplus \tilde{D}_{St} \}$ at the server sub-network to reduce the distance between the collection to the global feature distribution $\tilde{D}$. With this approach, SplitAVG lowers the upper bound of aggregated model risk $L_T (h)$ without touching raw data space $D_k$. It is worth mentioning that according to the bound, cut layer selection in SplitAVG does not affect generalization performance of the final model if $d(\tilde{D}_k, \tilde{D})$ is determined, instead, it affects model’s learning ability in extracting knowledge from $\tilde{D}$ and draw hypothesis. This influence might differ across medical imaging tasks but follow the same empirical conclusion that the earlier cut layer is, the more parameters server sub-network contains to facilitate learning.

\begin{figure*}[!t]
\centerline{\includegraphics[width=\columnwidth]{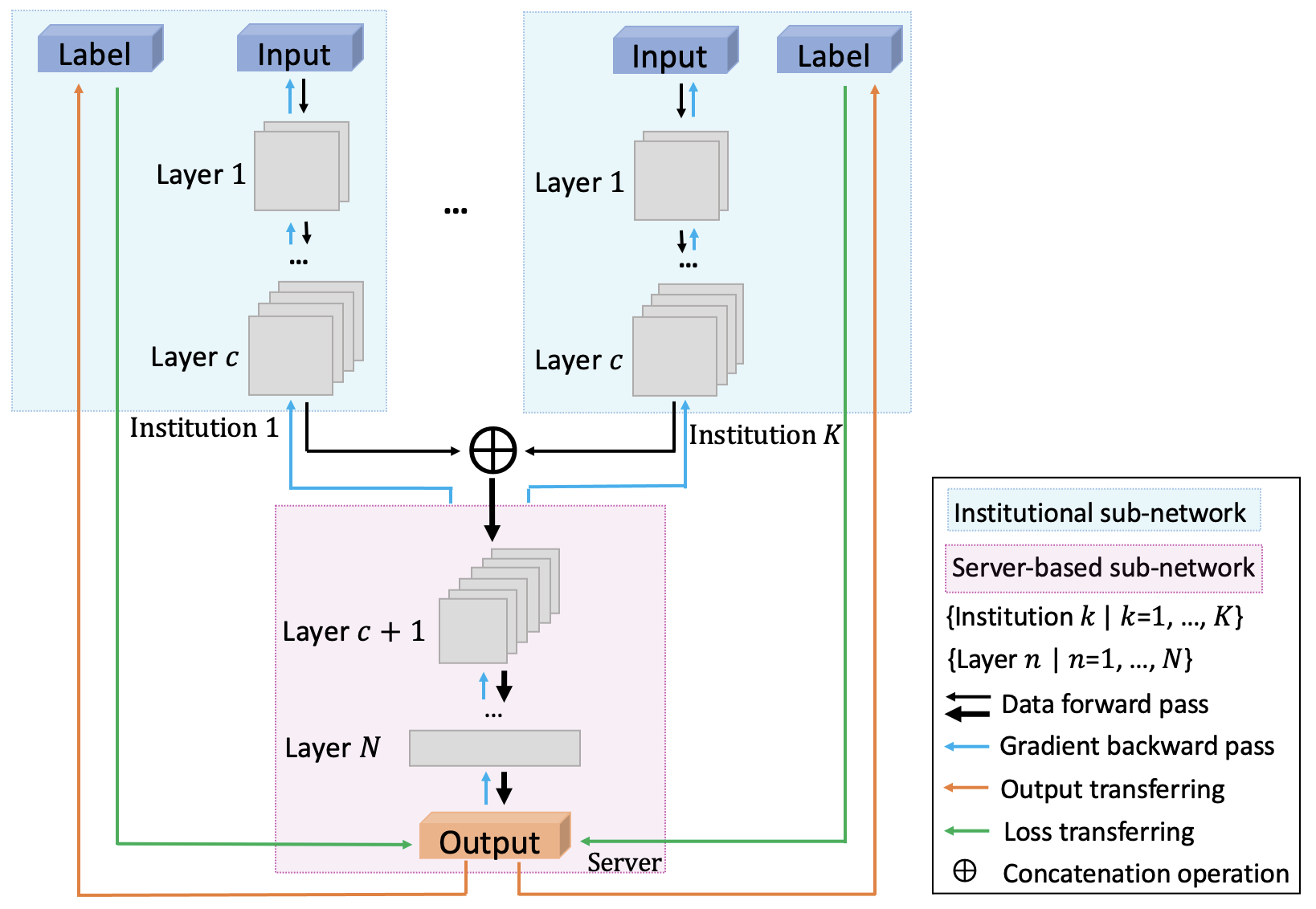}}
\caption{SplitAVG-v2: A variant of SplitAVG architecture which does not require local institutions sharing data labels to the central server.  }
\label{ArchitectureV2}
\end{figure*}

\subsection{SplitAVG-v2}
The proposed SplitAVG algorithm includes the process of local institutions sending data labels to the central server, which brings in risks of privacy leaking, especially in tasks with high-dimensional labels. To this end, we introduce SplitAVG-v2, an improved version of SplitAVG by keeping the labels in local institutions to solve the privacy leakage concern raised from label sharing. As the architecture shown in Figure ~\ref{ArchitectureV2}, we further introduce a split point in the later part of the server network, where output predictions are split into chunks that each chunk of predictions is derived from an institution’s data. Prediction chucks are sent back to corresponding institutions and a scalar loss is computed with local data labels. The server collects institutional losses and generates the final loss and gradients. 

SplitAVG-v2 does not require institutions to share raw data or raw labels, while retaining SplitAVG’s essence in generating unbiased gradients from collected loss. To illustrate, we take cross entropy loss as example. The traditional cross entropy in central server of SplitAVG is defined as:

\begin{equation}
\mathcal{L}_{CE} = -\sum_{i=1}^N[\sum_{c=1}^C t_{ic}log(p_ic)]
\end{equation}
, where $N$ is the number of data points, $C$ is is the number of classes, $t_{ic}$ is the true label and $p_{ic}$ is the SoftMax probability of class $c$ at data point $i$. 

In SplitAVG-v2, we defined an institutional cross-entropy:
\begin{equation}
\mathcal{L}_{CE_k} = -\sum_{i=1}^{N_k} \sum_{c=1}^C t_{ic}log(p_ic)
\end{equation}
, where $N_k$ is the number of data points at the $k^{th}$ client. $\mathcal{L}_{CE_k}$ is independently computed at each local client thus preserves the privacy of label $t_{ic}$. The server in SplitAVG-v2 then collects the institutional cross-entropy from all local institutions, resulting in the overall loss $L_{CEK}$ as: 
\begin{equation}
\mathcal{L}_{CEK} = \sum_{k=1}^K \mathcal{L}_{CE_k} = -[\sum_{k=1}^K \sum_{i=1}^{N_k}][\sum_{c=1}^C t_{ic}log(p_ic)] = \mathcal{L}_{CE}
\end{equation}
Even though a split point is introduced in SplitAVG-v2, the overall loss of SplitAVG-v2 is the same to the SplitAVG. Thus SplitAVG-v2 and SplitAVG will have the same experimental results if a same experimental setting is used for both models. We further tested SplitAVG-v2 on split 4 of Retina dataset, and obtained the identical results, mean accuracy of 76.5\%, with the SplitAVG result.

\subsection{Dataset and data partitions}

We evaluate our method on a set of both synthetic and real-world federated datasets, including the simulated federated datasets by artificially introducing data heterogeneity on a Diabetic Retinopathy (Retina) binary classification dataset \cite{kaggle} and a Bone Age (BoneAge) prediction dataset \cite{kaggleBone}, and the real-world federated Brain Tumor Segmentation (BraTS 2017) segmentation dataset \cite{Menze2015-kj, Bakas2017-hk, Carver2019-fy}:

The Retina dataset consists of 44 351 pairs of left and right eye color digital retinal fundus images obtained from the Kaggle Diabetic Retinopathy competition \cite{kaggle}. Each image is labeled on a scale of 0-4 based on the disease severity of diabetic retinopathy (DR), where 0 indicates no DR, and 1-4 represent mild, moderate, severe, and proliferative DR, respectively. We binarize the image labels to Healthy (scale 0) and Diseased (scale 2, 3 or 4) to simplify model training, and the mild DR (scale 1) images were excluded \cite{Chang2018-rg}. Furthermore, we only utilize left eye images to avoid the possible confusion from inconsistent correlation between disease presence in left/right eyes. The dataset is randomly sampled to create a training set of 6000 images, a validation set of 3000 images, and a testing set of 3000 images. The images are pre-processed following Ben Graham’s methods \cite{Graham2015-ou}: rescaled to a radius of 300, subtracting the local average color, image clipping for boundary removal, and resized to 256 x 256 resolution. Random cropping (to 224 x 224), random rotations (0, 90, 180, or 270 degrees) and horizontal flips were applied for data augmentation.

The BoneAge dataset consists of 14 236 pediatric hand radiographs obtained from the Kaggle Radiological Society of North America (RSNA) Bone Age competition \cite{kaggleBone}. Each image is labeled with the skeletal age provided by expert reviewers. The dataset is randomly sampled to create a training set of 4572 images, a validation set of 1000 images, and a testing set of 1000 images.

\begin{figure*}[h]
\centerline{\includegraphics[width=7in]{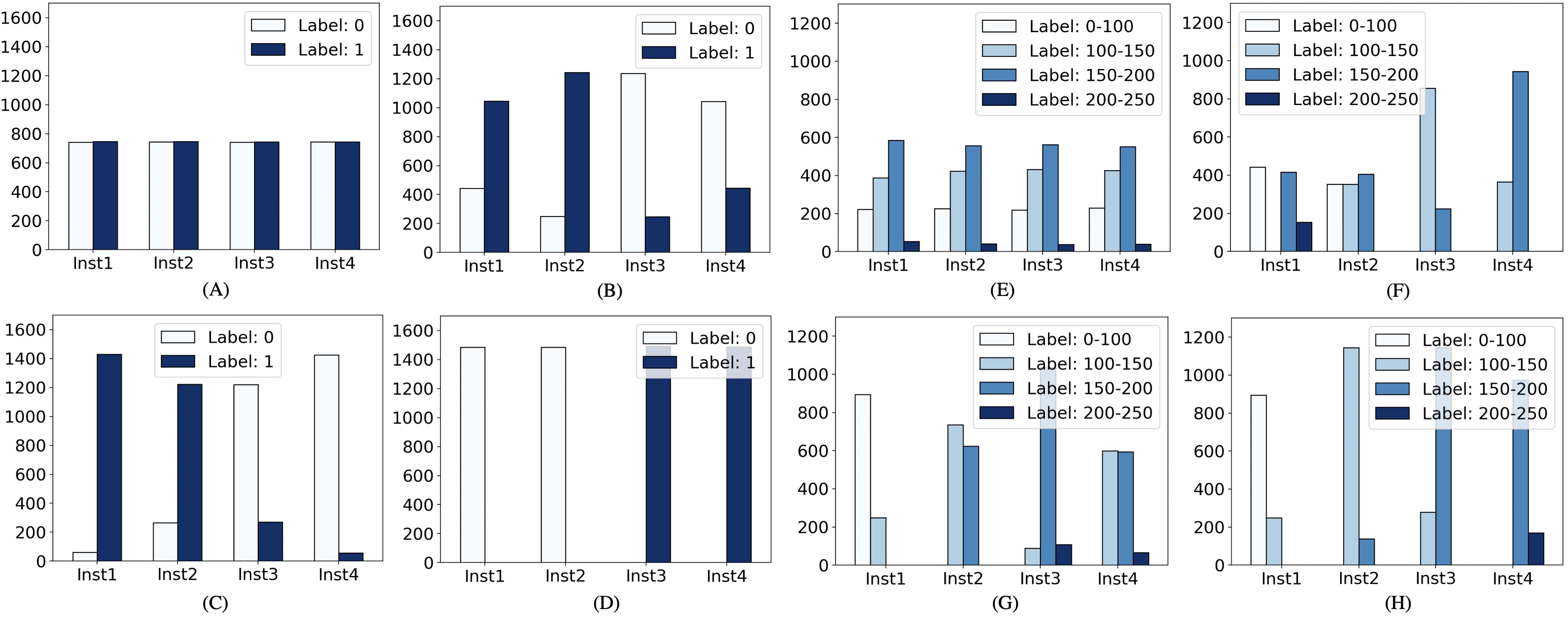}}
\caption{Simulated data partitions on Retina and BoneAge datasets to simulate heterogeneity in data among 4 simulated institutions. Data partitions on Retina dataset with (A) KS=0, (B) KS=0.40, (C) KS=0.56 and (D) KS=0.67. Data partitions on BoneAge dataset with (E) KS=0.29, (F) KS=0.59, (G) KS=0.73, (H) KS=0.97.}
\label{Distribution}
\end{figure*}

We simulate four “institutions” and create four kinds of data partitions for both the Retina and BoneAge datasets: one homogenous data partition, and three heterogeneous data partitions with label distribution skew. The degree of label distribution skew is controlled by the fraction of non-IID data and is scaled by the mean Kolmogorov-Smirnov (K-S) statistic between every two institutions. Specifically, the K-S value being 0 indicates homogeneity and 1 indicates entirely different distributions. Fig.~\ref{Distribution}  depicts the detailed data partitions.

The BraTS dataset consists of magnetic resonance imaging (MRI) brain scans of gliomas collected from multiple institutions\cite{Menze2015-kj, Bakas2017-hk, Carver2019-fy}. Each scan is manually labeled with segmentation annotations of tumor regions \cite{Menze2015-kj, Bakas2017-hk, Carver2019-fy}. In our experiments, we focus on the segmentation for whole tumor region, and we only use high-grade glioblastoma (HGG) scans in T2 Fluid Attenuated Inversion Recovery (FLAIR) modality. We randomly select scans from 45 subjects as the testing set and the rest scans (120 subjects) as the training set. As a real-world federated dataset, BraTS includes common types of data heterogeneity, i.e., imaging acquisition skew (the scans
are collected from ten institutions with different imaging equipment and protocols), label distribution skew and sample size distribution skew (one institution contributes 69 subjects while some institutions only contribute 4 or 5 subjects).

\subsection{Comparison methods }

We compare our SplitAVG method with seven state-of-the-art federated learning methods including four traditional methods: FedAvg \cite{McMahan2017-ip}, FedSGD \cite{McMahan2017-ip}, CWT \cite{Chang2018-rg}, and SplitNN \cite{Gupta2018-dq}, and three optimized methods proposed for non-IID data: Federated stochastic gradient descent with group normalization (FedSGD+GN) \cite{Hsieh2020-wx}, Federated averaging with server momentum (FedAvgM) \cite{Hsu2019-sd}, and Federated averaging with globally shared data (FedAvg+SD) \cite{Zhao2018-kq}. We use the performance of a model trained with centrally hosted data as the baseline approach, termed as “centrally hosted”. This represents the ideal situation for training deep learning models since all data are centralized.

\textbf{FedAvg} is an aggregation-based method. For each epoch, local institutions conduct $\frac{Q_k}{B}$ training iterations, then transfer model weights to a central server, which averages the weights and transfers the updated weights back to individual institutions \cite{McMahan2017-ip}. $Q_k$ is the quantity of training samples at institution $k$, and $B$ is the local mini-batch size.

\textbf{FedSGD} is a full-communication version of FedAvg. For each training iteration, local institutions transfer model gradients to a central server, which generates weights updated from aggregated gradients and transfers the updated weights back to individual institutions \cite{McMahan2017-ip}.

\textbf{CWT} is a transfer-based method. For each epoch, local institutions conduct $\frac{Q}{B \times K}$ training iterations, where $Q$ is the quantity of training samples of the centrally hosted data, and cyclically transfers model weights to the next training institution until model convergence \cite{Chang2018-rg}.

\textbf{SplitNN} is a transfer-based method. For each epoch, local institutions conduct $\frac{Q}{B \times K}$ training iterations with weights and gradients transferred between institutions and the server. Specifically, for each iteration: (1) a local institution forward propagates training data until the cut layer and transfers the outputs at the cut layer to a central server, (2) the server completes the rest of the training with the received output, (3) the server generates gradients, back propagates through the cut layer to the institution, and updates the model weights, and (4) the institution transfers model weights to the next training institution \cite{Gupta2018-dq}. Similar to our SplitAVG, SplitNN also splits the whole network architecture into two parts, and involves frequent transfer of intermediate feature maps and gradients between the central server and local institutions. However, unlike SplitAVG that trains local institutional sub-networks in parallel and uses an aggregation operation to concatenate the intermediate feature maps in the central server, SplitNN directly uses a serial and cyclical transfer training mode in each local institution, which always suffers from catastrophic forgetting when data heterogeneity exists across institutions.

\textbf{FedSGD+GD} is an optimization method for FedSGD, which applies GroupNorm layers to avoid the skew-induced accuracy loss of batch normalization layer for non-IID data \cite{Hsieh2020-wx}. We set the number of groups in GroupNorm layers to 32.

\textbf{FedAvgM} is an optimization method for FedAvg, which applies a momentum optimizer on the server to improve its robustness on non-IID data partitions \cite{McMahan2017-ip, Hsu2019-sd}. We set the momentum parameter to 0.9.

\textbf{FedAvg+SD} is an optimization method for FedAvg, which applies a data-sharing strategy to improve the training of FedAvg on non-IID data partitions \cite{McMahan2017-ip, Zhao2018-kq}. Specifically, 5\% of the global data was distributed and globally shared between all local institutions.

\subsection{Experimental setup}

We choose the 34-layer residual network (ResNet34) pre-trained on ImageNet as the base network for all methods on Retina and BoneAge dataset \cite{He2016-lw, Russakovsky2015-sg}. All methods are implemented in Pytorch and optimized using SGD \cite{ResNet}. The objective function for Retina classification task and BoneAge regression task is binary cross-entropy and L1-norm, respectively. We set the mini-batch size B to 32, the learning rate to 0.001 (scaled 0.1 every 40 epochs), the momentum coefficient to 0.9. Final models are evaluated by calculating the accuracy of testing data for the Retina classification task, and the mean absolute error (MAE) between true age values and predicting age values of testing data for the BoneAge regression task.

For BraTS segmentation task, we use U-Net as the base model and Dice Loss as the objective function \cite{U-Net, Dice1945-xh}. The final models are evaluated by Dice Similarity Coefficient (DSC) between the true and predicted boundaries \cite{Dice1945-xh}.

The numbers of selected institutions involved in each round of federated learning for all the comparing methods are set to 4 ($St=4$, $K=4$) for the synthetic datasets (Retina and BoneAge). To ensure that the communication and storage costs of SplitAVG does not increase with more institutions, we also set ($St=4$, $K=10$) for BraTS dataset.

\section{Results}

\begin{figure}[h]
	\centering
	\begin{center}
		\begin{tabular}{ccc}
	\includegraphics[width=0.9\linewidth]{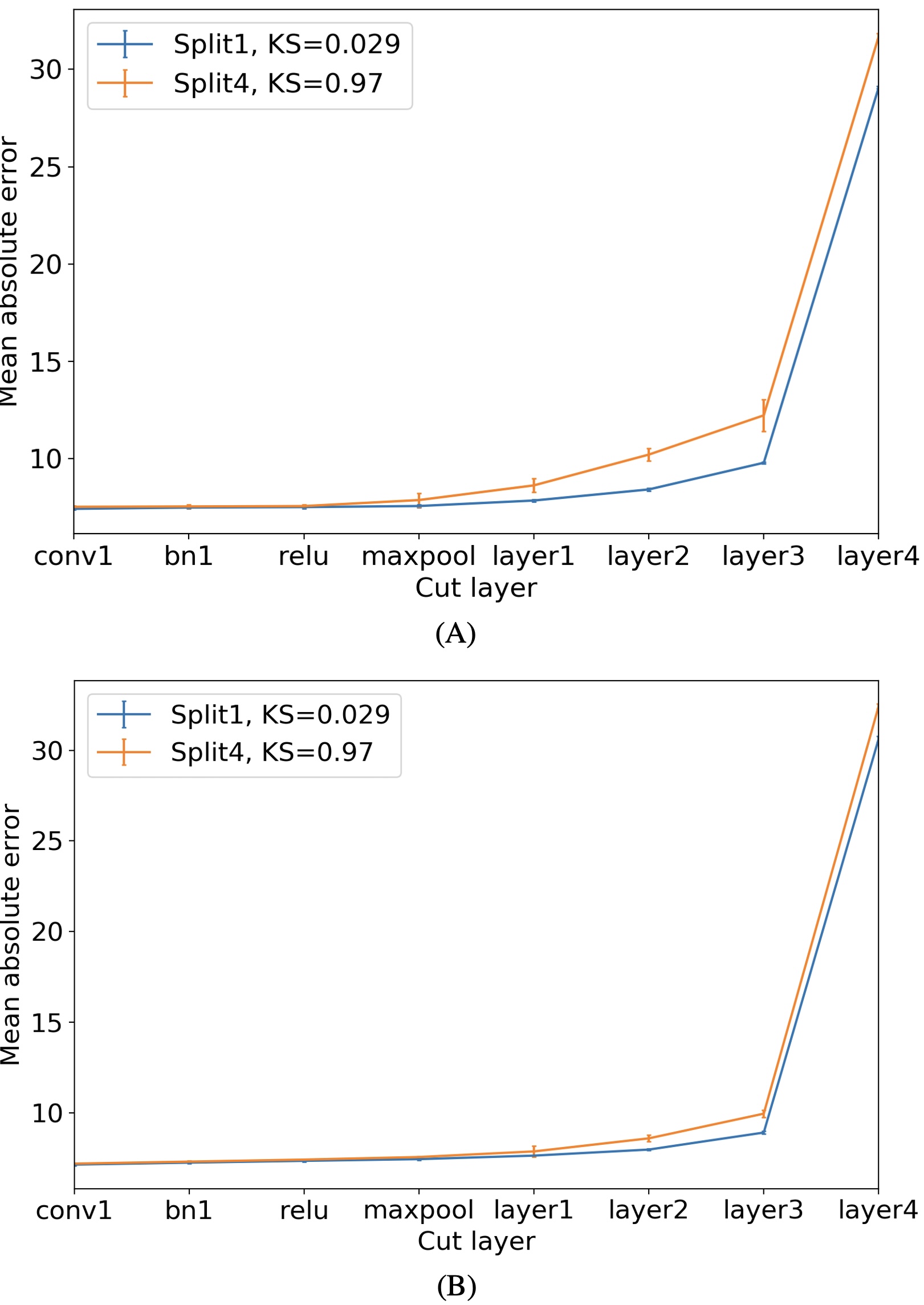}&
		\end{tabular}
	\end{center}
	\caption{For SplitAVG, when trained with different cut layers of ResNet34 (A) and ResNet50 (B) on a homogeneous split and a heterogeneous split of BoneAge dataset, the model performance on validation dataset is shown by the test mean absolute error (MAE).}
	\label{CutLayerSelection}
\end{figure}

\subsection{Cut layer selection for SplitAVG}

The selection of the affiliated cut layer for the institutional sub-network and server-based sub-network in SplitAVG affects the final model performance and needs to be carefully selected. We investigate the optimal cut layer for the base model ResNet34 and ResNet50 on BoneAge dataset with 1 homogeneous data partition and 1 heterogeneous data partition among 4 participating institutions. ResNet34 and ResNet50 consist of the following sequential layers: {“conv1”, “bn1”, “relu”, “maxpool”, “layer1”, “layer2”, “layer3”, “layer4”, “avgpool”, and “fc”}, which are tested as cut layers respectively \cite{ResNet}. The selection of cut layer affects the final model performance. However, the optimal cut layer selection is irrelevant to data partition types or base network types, that earlier cut layers tend to result in better performance (Fig.~\ref{CutLayerSelection}). This observation is consistent with what we inferred in theory analysis that SplitAVG method can learn more abundant unbiased information when the feature maps from local institutions are concatenated at earlier layers. Also, the results showed that deeper cut layers do not significantly compromise the model performance, especially as the base model complexity increases, comparing ResNet50 to ResNet34 results. The models fail only when setting last layer as cut layer, when the server network does not have sufficient model learnable parameters to interpret concatenated feature maps, so it can’t make valid predictions. Experimental results show that the ResNet34 model trained with “conv1” as the cut layer obtained a good performance in all settings. Therefore, we set “conv1” as the cut layer of ResNet34 for SplitAVG in all remaining experiments.

\begin{figure*}[h]
\centerline{\includegraphics[width=5.5in]{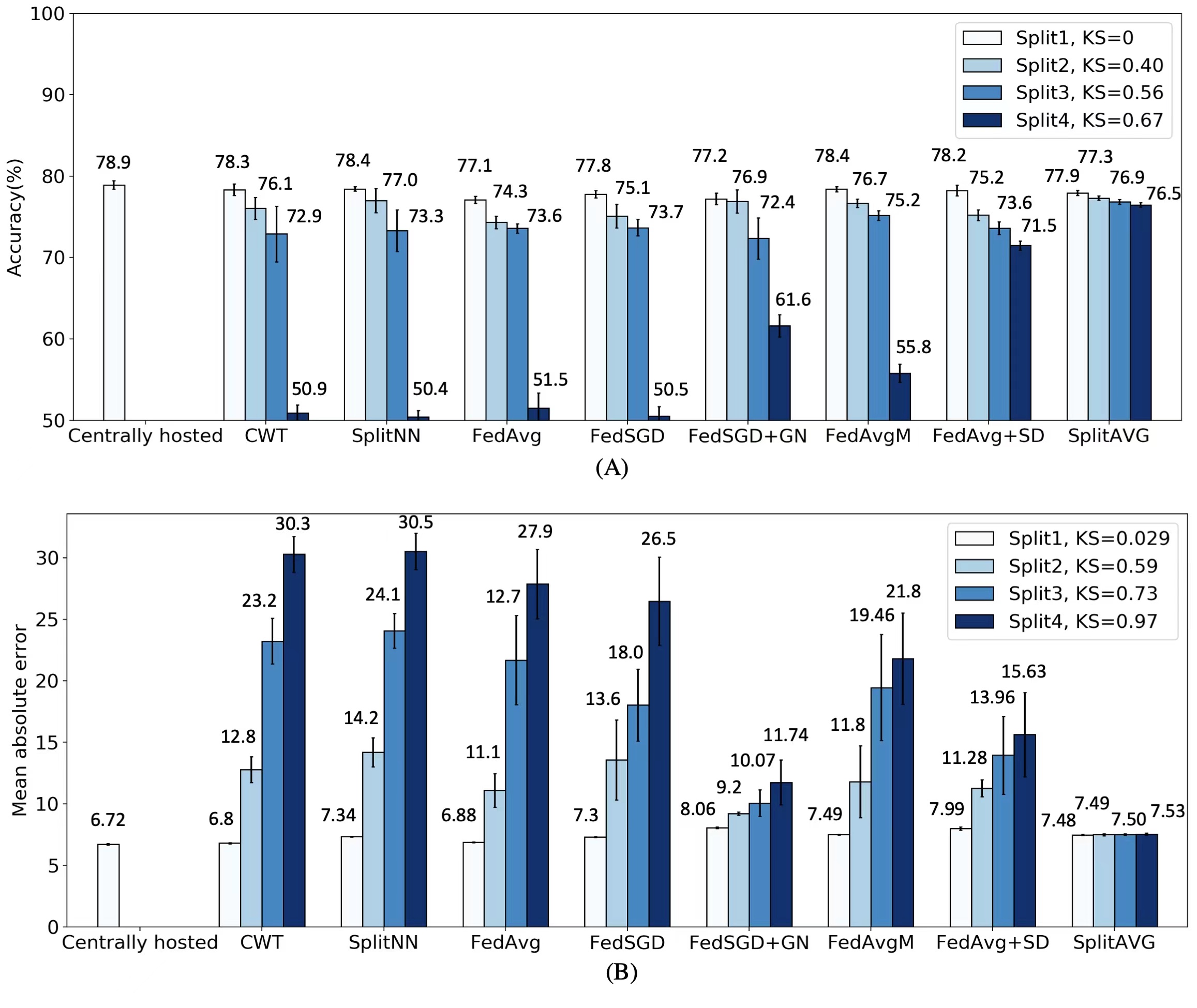}}
\caption{(A) The test accuracy on Retina splits and (B) the test mean absolute error (MAE) on BoneAge splits of all comparison methods.}
\label{MainResults}
\end{figure*}

\subsection{Model performance on synthetic federated datasets}

We evaluate the performance of SplitAVG on the Retina and BoneAge datasets with both homogeneous and heterogeneous data partitions (Splits 1-4 shown in Fig.~\ref{Distribution}) and compare it to seven state-of-the-art federated learning methods (FedAvg \cite{McMahan2017-ip}, FedSGD \cite{McMahan2017-ip}, CWT \cite{Chang2018-rg}, SplitNN \cite{Gupta2018-dq}, FedSGD+GN \cite{Hsieh2020-wx}, FedAvgM \cite{Hsu2019-sd}, FedAvg+SD \cite{Zhao2018-kq}). Fig.~\ref{MainResults} shows that all the compared federated methods perform well on the homogeneous data partition (Split 1) but lose significant accuracy on splits with label distribution skew (Splits 2-4) (Fig.~\ref{MainResults}). For example, CWT, SplitNN, FedAvg, and FedSGD lose 35.0\%, 35.7\%, 33.2\%, and 35.07\% prediction accuracy on Split 4 of the Retina dataset, respectively (Fig.~\ref{MainResults}(A)). The three optimized methods, FedSGD+GN, FedAvgM, and FedAvg+SD may help mitigate the performance loss for data partitions with mildly skewed label distributions, but still diverge severely on splits with highly skewed label distributions. For example, even when 5\% of centrally hosted data are globally shared among each institution, the prediction accuracy of FedAvg+SD is 8.6\% lower on Split 4 (KS=0.67) than that on homogenous Split 1 of Retina dataset (Fig.~\ref{MainResults}(A)). For SplitAVG, there is only 1.89\% drop in accuracy on Split 4 of Retina data (Fig.~\ref{MainResults}(A)), and 0.733\% MAE rise on Split 4 of BoneAge data (Fig.~\ref{MainResults}(B)), than that on homogenous Split 1. In each training iteration, FedSGD method transfers $2.13 \times 10^7$ data (as float32) from local institutional model to the server, while SplitAVG only requires $8.03 \times 10^5 $ data transfer.

\begin{figure}
\centerline{\includegraphics[width=0.999\columnwidth]{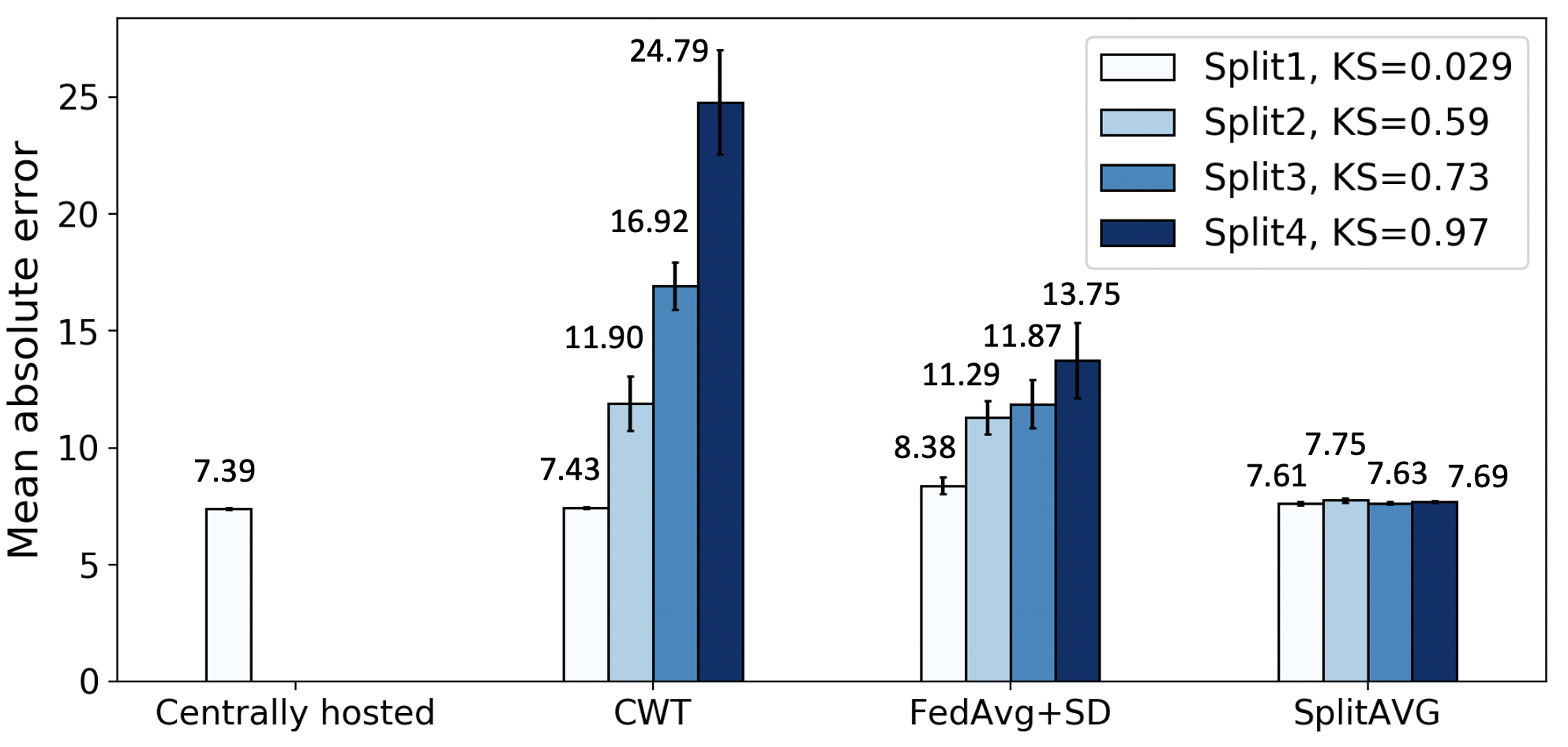}}
\caption{The test mean absolute error (MAE) of CWT, FedAvg+SD, and SplitAVG on BoneAge dataset splits when MobileNet-v2 is applied as the base model.}
\label{BoneAgeMobileNet}
\end{figure}

\subsection{Model robustness for a different deep learning architecture}

We replace the base model ResNet34 with MobileNet-v2 for all algorithms \cite{Sandler2018-ct}. The architecture of MobileNet-v2 consists of a “Feature” structure including 16 “InvertedResidual” blocks, and a “Classifier” layer \cite{MobileNet}. Following the cut layer empirical results in ResNet34, we set the first “InvertedResidual” block in MobileNet-V2 as the cut layer. The predicted MAE is used to evaluate SplitAVG, CWT, and FedAvg+SD on the BoneAge dataset when MobileNet-v2 is used as the base model. SplitAVG again demonstrates the best performance among all the compared methods. On the most skewed data partition (Split 4), SplitAVG achieves 104.7\% of the MAE obtained by the baseline (Fig. \ref{BoneAgeMobileNet}).

\begin{figure*}[h]
\centering
    \includegraphics[width=1.6\columnwidth]{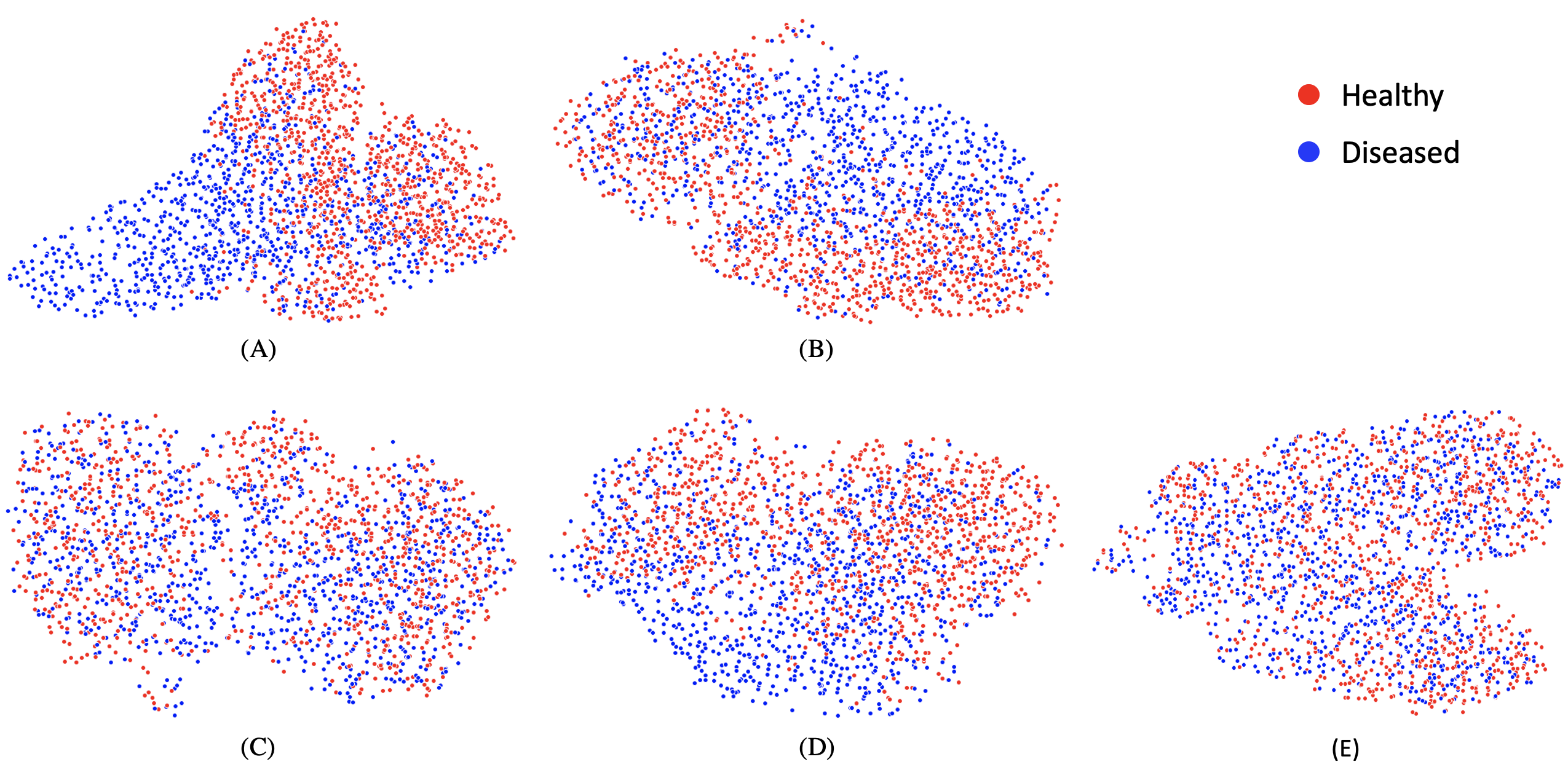}
    \caption{Feature embedding visualization of (A) baseline centrally hosted, (B) SplitAVG, (C) FedAvgM, (D) FedAvg+SD, (E) FedSGD + GN on highly heterogeneous data splits (KS=0.67) of Retina dataset using UMAPs. Here, ResNet34 is applied as the base network.}
    \label{Umap}
\end{figure*}
\subsection{SplitAVG on a real-world federated dataset }

\begin{figure}[h]
\centerline{\includegraphics[width=\columnwidth]{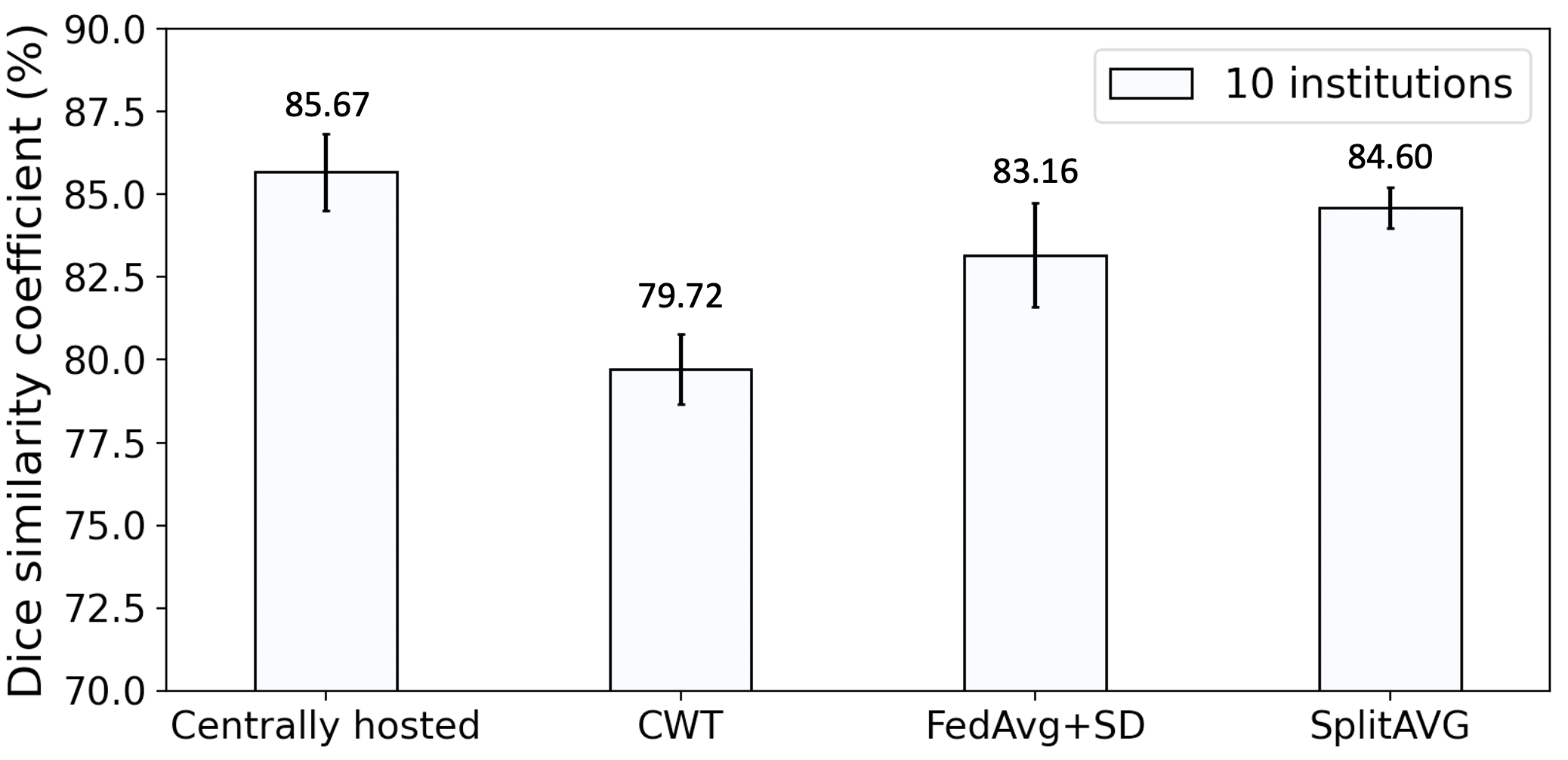}}
\caption{The dice similarity coefficient (DSC) of centrally hosted baseline, CWT, FedAvg+SD, and SplitAVG on BraTs dataset when U-Net is applied as the base model.}
\label{BratsResult}
\end{figure}

\begin{figure}[h]
\centerline{\includegraphics[width=\columnwidth]{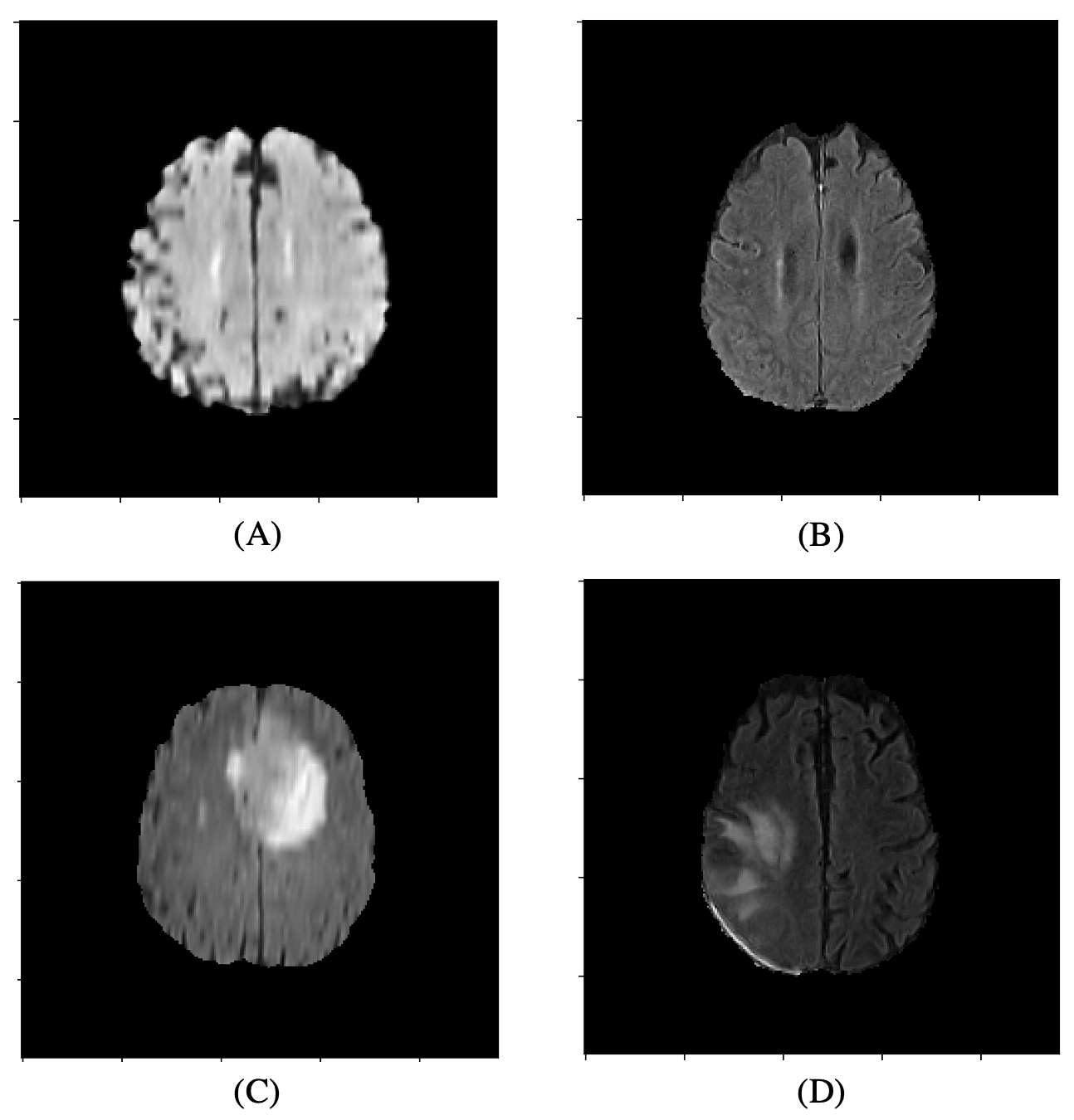}}
\caption{Examples of images (with varying intensity, image contrast, and etc) obtained from different institutions of BraTs dataset.}
\label{Brain}
\end{figure}

We use the BraTS segmentation dataset to test SplitAVG’s robustness to the real-world data heterogeneity settings \cite{Menze2015-kj, Bakas2017-hk, Carver2019-fy}. The BraTs dataset contains multi-modal magnetic resonance imaging (MRI) scans of 285 subjects with brain tumors. It is collected from 10 institutions with varying equipment and imaging protocols, thus resulting in heterogeneous data distributions among different clients, see Fig. \ref{Brain} for four examples of images obtained from different institutions. Following \cite{Sheller2019-nj}, we test the performance of SplitAVG on the whole tumor volume segmentation task and adopt the FLAIR modality as the input, comparing with the centrally hosted baseline, CWT, and FedAvg+SD methods. We performed three trials with each method and take the mean of segmentation results across the 10 participating institutions. The model trained with data centrally hosted obtained the mean DSC result of 85.67\%, and the model trained with CWT, FedAvg+SD, and SplitAVG obtained the mean DSC results of 79.72\%, 83.16\%, and 84.6\%, as shown in Fig. \ref{BratsResult}.

\subsection{Analyzing SplitAVG from interpretation perspective}

We further visualize the latent space embedding of the features (the first “fc” layer of ResNet34) from the models trained with our SplitAVG, three federated learning optimization methods, and the baseline centrally hosted training, to aid our understanding of different model’s robustness on heterogeneous data splits from interpretation perspective. We use Retina test dataset and draw features computed over samples of healthy label and diseased label with two different colors with UMAPs \cite{Sheller2019-nj}. As shown in Fig.~\ref{Umap}, the baseline UMAP presents the best clustering for same classes. Among UMAPs of comparing federated learning methods, SplitAVG shows the clearest separation for different classes, while the features of healthy and diseased shown from FedAvgM, FedAvg+SD, and FedSVG+GN are highly entangled. This experiment again demonstrates the superiority of SplitAVG on heterogenous data.

\section{Discussion}

Federated learning has emerged as an attractive paradigm for enabling collaboratively training deep learning models without sharing patient data. Although numerous federated learning approaches have been proposed, a critical aspect of existing federated learning methods is that they either assume the data are IID across institutions or only consider mild skewed non-IID data distribution. The performance of models trained using these federated learning methods degrades with increasing degrees of data heterogeneity. In this study, we develop a heterogeneity-aware optimization platform, SplitAVG, to address the challenge of data heterogeneity in federated learning methods.

We first evaluate our SplitAVG method on the simulated distributed data by artificially introducing various degrees of label distribution skew on the Retina binary classification dataset \cite{kaggle} and bone age prediction dataset \cite{kaggleBone} and compare it with seven state-of-the-art federated learning methods. We found that all the compared federated learning methods are vulnerable to label distribution skew. For Retina test dataset, the accuracy of models trained using FedAvg, FedSGD, CWT, and SplitNN, decreases from 74.3\%, 77.8\%, 78.3\%, and 78.4\% on data partitions with mild degree of label distribution skewness (KS=0.40) to 51.5\%, 50.5\%, 50.9\%, and 50.4\% on data partitions with high degree of label distribution skewness (KS=0.67), respectively. Even with complex heuristic parameters tuning (e.g., FedSGD+GN requires the extra pre-training on the model with GN layers \cite{Hsieh2020-wx}, and FedAvgM includes the tuning for momentum parameters \cite{Hsu2019-sd}) or with the risk of sharing partial raw data (FedAvg+SD) \cite{Zhao2018-kq}, the compared methods still suffer from severe performance drops on highly heterogeneous data partition. With the help of simple network splitting strategy and the concatenation operation of intermediate feature maps, our SplitAVG, however, successfully mitigates model performance loss caused by the label distribution skew even in the extreme heterogeneous cases.

We then investigate whether SplitAVG method can handle other kinds of data heterogeneity besides label distribution skew and if it performs well in other kinds of deep learning tasks besides image classification and regression. Experimental results on a real-world BraTS segmentation dataset show that, even when tested with a mixture of various types of data heterogeneity (quantity skew, imaging acquisition skew, label distribution skew, etc), SplitAVG still achieves comparable performance to the baseline centrally hosted case. In contrast to previous methods that different optimizations are required for each type of data heterogeneity, for example cyclic weighted loss for tackling label heterogeneity and proportional local training for handling sample size heterogeneity \cite{Balachandar2020-go}, our SplitAVG is more scalable and that can more broadly address the challenge of data heterogeneity across centers in federated learning.

One limitation of our work is that we only study the performance of federated learning methods with statistical data heterogeneity. There are other sources of heterogeneity, such as device heterogeneity (e.g., variation in computer hardware's and communication speeds) and behavior heterogeneity (e.g., institutions may join in or drop out during federated model training at any time), which is an important area for future work. One data privacy concern for the proposed SplitAVG is the risk of reconstructing raw images from shared feature maps of the cut layer, which can be prevented by integrating privacy protecting techniques such as secure multi-party computation (MPC) \cite{Goldreich1998-kb} and differential privacy \cite{abadi2016deep}, and future work can develop adjusted configurations for SplitAVG combining these techniques.

\section{Conclusion}
In this paper, we have proposed SplitAVG, a heterogeneity-aware optimization platform that tackles fundamental and pervasive data heterogeneity problems inherent in federated learning. Our SplitAVG can be consumed as an off-the-shelf federated learning platform and provides immediate improvements, without any complex hyper-parameter tuning, training heuristic, or additional training/fine-tuning. Our SplitAVG is also model agnostic and can be generalized to various types of medical imaging tasks. Experimental evaluation of SplitAVG on a suite of both simulated and real-world federated datasets with various degrees of non-IID data partitions, and its comparisons with seven state-of-the-art federated learning methods and a baseline of centrally hosted data demonstrate the effectiveness of SplitAVG method in handling common types of heterogeneous data across institutions. The findings in this work provide a promising solution to overcoming the challenge of heterogeneous data in real-world federated learning settings.


\end{document}